\PassOptionsToPackage{unicode}{hyperref}
\PassOptionsToPackage{hyphens}{url}
\documentclass[
]{article}
\usepackage{xcolor}
\usepackage{amsmath,amssymb}
\usepackage{iftex}
\ifPDFTeX
  \usepackage[T1]{fontenc}
  \usepackage[utf8]{inputenc}
  \usepackage{textcomp} % provide euro and other symbols
\else % if luatex or xetex
  \usepackage{unicode-math} % this also loads fontspec
  \defaultfontfeatures{Scale=MatchLowercase}
  \defaultfontfeatures[\rmfamily]{Ligatures=TeX,Scale=1}
\fi
\usepackage{lmodern}
\ifPDFTeX\else
\fi
\IfFileExists{upquote.sty}{\usepackage{upquote}}{}
\IfFileExists{microtype.sty}{% use microtype if available
  \usepackage[]{microtype}
  \UseMicrotypeSet[protrusion]{basicmath} % disable protrusion for tt fonts
}{}
\makeatletter
\@ifundefined{KOMAClassName}{% if non-KOMA class
  \IfFileExists{parskip.sty}{%
    \usepackage{parskip}
  }{% else
    \setlength{\parindent}{0pt}
    \setlength{\parskip}{6pt plus 2pt minus 1pt}}
}{% if KOMA class
  \KOMAoptions{parskip=half}}
\makeatother
\usepackage{longtable,booktabs,array}
\usepackage{caption}
\usepackage{multirow}
\usepackage{calc} % for calculating minipage widths
\usepackage{etoolbox}
\makeatletter
\patchcmd\longtable{\par}{\if@noskipsec\mbox{}\fi\par}{}{}
\makeatother
\IfFileExists{footnotehyper.sty}{\usepackage{footnotehyper}}{\usepackage{footnote}}
\makesavenoteenv{longtable}
\usepackage{graphicx}
\makeatletter
\newsavebox\pandoc@box
\newcommand*\pandocbounded[1]{% scales image to fit in text height/width
  \sbox\pandoc@box{#1}%
  \Gscale@div\@tempa{\textheight}{\dimexpr\ht\pandoc@box+\dp\pandoc@box\relax}%
  \Gscale@div\@tempb{\linewidth}{\wd\pandoc@box}%
  \ifdim\@tempb\p@<\@tempa\p@\let\@tempa\@tempb\fi% select the smaller of both
  \ifdim\@tempa\p@<\p@\scalebox{\@tempa}{\usebox\pandoc@box}%
  \else\usebox{\pandoc@box}%
  \fi%
}
\def\fps@figure{htbp}
\makeatother
\usepackage{bookmark}
\IfFileExists{xurl.sty}{\usepackage{xurl}}{} % add URL line breaks if available
\makeatletter
\@ifundefined{xmpquote}{}{}
\makeatother
\hypersetup{
  hidelinks,
  pdfcreator={LaTeX via pandoc}}

\title{A Dual-Dimensional LLM Framework for Automated Item Incidental Content Similarity Analysis in Large-Scale Assessments}
\author{%
  Jing Huang\textsuperscript{a*}, Jihong Zhang\textsuperscript{b}, Hua-Hua Chang\textsuperscript{a}\\[4pt]
  \small \textsuperscript{a}College of Education, Purdue University, West Lafayette, USA\\
  \small \textsuperscript{b}Educational Statistics \& Research Methods, University of Arkansas, Fayetteville, USA\\[4pt]
  \small *Corresponding author: \href{mailto:huan1936@purdue.edu}{\nolinkurl{huan1936@purdue.edu}}
}
\date{}

\begin{document}

\maketitle

\textbf{Abstract.} The rapid expansion of large-scale assessments and the growing adoption of automatic item generation have intensified concerns about incidental content redundancy, where construct-irrelevant elements such as wording or contextual framing become unintentionally repetitive across items. Traditional similarity metrics like BLEU or cosine similarity, often fail to capture the nuanced structural and semantic layers that drive perceived redundancy simultaneously. This study proposes a dual-dimensional framework for Automated Item Similarity Analysis (AISA) powered by Large Language Models (LLMs), operationalizing similarity through Structured Decomposition and Semantic Relatedness. Psychometric validation indicates that LLM-derived metrics align more closely with indicators of construct-irrelevant local dependence and yield more coherent item parameter groupings than traditional text-based measures. The framework is further evaluated through its application in Computerized Adaptive Testing (CAT). Simulations reveal that incorporating LLM-based similarity constraints into item selection improves estimation stability and reduces bias with minimal efficiency trade-offs, outperforming constraints based on conventional metrics. These findings highlight the potential of LLM-powered AISA to support scalable bank curation, content-aware test assembly, and experience-sensitive adaptive testing across diverse assessment contexts.

\textbf{Keywords:} Generative AI, Semantic Proximity, Item Cloning, Test-Taker Experience

\section{\texorpdfstring{\textbf{Introduction}}{Introduction}}\label{introduction}

The integrity of educational and psychological assessments depends fundamentally on the quality and diversity of the underlying item banks. As the demand for large-scale, continuous testing grows, the field is experiencing a paradigm shift in item development, moving from traditional, expert-based authoring to Automatic Item Generation (AIG) supported by large language models (LLM) and generative AI (Götz et al., 2023; Song et al., 2025). While AIG has successfully addressed the challenge of item quantity, it has simultaneously introduced new challenges for quality assurance. One of these considerations is mitigating incidental content redundancy, which refers to a phenomenon where construct-irrelevant elements of item (e.g., scenarios, phrasings, linguistic structures; Ford et al., 2000), become unexpectedly repetitive across items. Incidental content redundancy often goes overlooked compared to construct targeted content, however, it may also create tangible risks in operational assessment, such as practice effects and test-taking fatigue (Wise \& Kingsbury, 2022). Over time, such incidental redundancy may ultimately threaten the accuracy of score interpretations (Haroz et al., 2020; Wise \& Kingsbury, 2022). Consequently, a robust, automated method for item similarity analysis is no longer a luxury but a necessity for modern item bank curation and test delivery.

However, existing automated methods for quantifying text similarity are fragmented in their capabilities and cannot holistically assess incidental content. Widely used automated text similarity metrics, such as BLEU (Bilingual Evaluation Understudy; Papineni et al., 2002), ROUGE (Recall-Oriented Understudy for Gisting Evaluation; Lin, 2004), and METEOR (Metric for Evaluation of Translation with Explicit Ordering; Lavie \& Denkowski, 2009), have proven useful for capturing lexical overlap, but often struggle to represent deeper semantic relationships (Rodriguez-Torrealba et al., 2022; Xu et al., 2024; Li et al., 2024). On the other hand, more context-based metrics, including cosine similarity, unit overlap, and longest common subsequence (Rodriguez-Torrealba et al., 2022; Kilmen \& Bulut, 2025), can better capture general lexical semantics, but they still operate on token-level or vector representations that often conflate stylistic features with substantive meaning and lack sensitivity to logical item structure (Steck et al., 2024; Zhou et al., 2022). Thus, no single prevailing approach can simultaneously evaluate both the structural and semantic dimensions essential for a complete understanding of incidental redundancy.

Recent advances in LLMs offer a compelling opportunity to overcome these limitations. Pretrained on vast corpora, LLMs possess a profound capacity for contextual understanding, enabling them to disentangle surface form from deep meaning, and to evaluate semantic coherence and logical structure in a scalable, reproducible manner (Li et al., 2024; Butterfuss \& Doran, 2025). These advantages make LLMs a powerful alternative to human reviewers and ideal engines for next-generation, comprehensive item similarity analysis.

Building on these advancements, this research proposes an LLM-powered item similarity analysis framework that quantifies item similarity through integrated structural and semantic analysis. The framework is intended as a general-purpose tool for modern assessment contexts in which large item banks are developed, curated, and deployed at scale, including automatic item generation quality control, item pool diagnostics, and test assembly. To demonstrate its operational value in a high-stakes and methodologically demanding setting, this study applies the framework within Computerized Adaptive Testing (CAT). Existing CAT item selection mechanisms, such as content balancing and exposure control, provide only partial safeguards, as they rely on predefined categorical labels or usage frequency without regard to the actual semantic relationships between items (Huang et al., 2025). By integrating LLM-derived similarity information into a cluster-constrained item selection algorithm, this study moves beyond traditional content balancing and illustrates how automated similarity analysis can support the joint goals of psychometric precision and content diversity. This application highlights the broader potential of the framework to enhance test quality and examinee experience across diverse assessment systems.

\section{\texorpdfstring{\textbf{Background}}{Background}}\label{background}

\subsection{\texorpdfstring{\emph{\textbf{2.1 Incidental Content and Its Implications for Computerized Adaptive Testing}}}{2.1 Incidental Content and Its Implications for Computerized Adaptive Testing}}\label{incidental-content-and-its-implications-for-computerized-adaptive-testing}

In traditional item development process, subject matter experts (SMEs) are trained to author items with attention to three types of content (Ford et al., 2000). \emph{\textbf{Targeted content}} refers to the knowledge or skills the test is designed to measure, typically evaluated using psychometric techniques such as structural equation modeling (SEM; Ding \& Hershberger, 2002) or exploratory factor analysis (EFA; Beavers et al., 2013). \emph{\textbf{Inappropriate content}} introduces construct-irrelevant variance, such as cultural or gender bias, and is commonly assessed using differential item functioning (DIF) analyses (Raju et al., 1995).

The third dimension, \emph{\textbf{incidental content}}, has historically received less systematic attention. According to Ford et al. (2000), this dimension shapes the presentation of the item rather than its core construct and is needed as supporting details for scenario and real-world application items. Sonnleitner et al. (2025) described the incidentals as the aspects that do not affect item difficulty and can be varied across items. To operationalize incidental content for systematic analysis, it is essential to recognize that it is not monolithic but rather comprises multiple, distinguishable dimensions. When examining the relationship between task features and item difficulties, previous studies used the lexical and semantic similarity to present the correspondence effects (Gorin \& Embretson, 2006). This multifaceted view is further supported by research across assessment domains. In automated essay scoring for L2 writing, Hur and Ji (2025) explicitly structured their analysis around two complementary dimensions: linguistic features (e.g., lexical diversity, syntactic complexity) assessing writing form, and a semantic relevance feature assessing writing content. Similarly, in reading comprehension, Sayin and Gierl (2024) also explicitly identified distinct presentational features, including semantic features (e.g., text meaning), organization features (e.g., text structure and order), and textual features (e.g., word count and text readability). Building on this convergent evidence, this study conceptualizes \emph{\textbf{incidental content}} as encompassing both semantic (e.g., real-life scenarios) and linguistic/structural features (e.g., item format, sentence structures, and language complexity).

To clarify the distinction between the targeted and incidental content, and to further illustrate the importance of incidental content, consider a simple math item like 2 + 2 = ?, originally discussed by Ford et al. (2000). When contextualized as, ``\emph{Jackson received 2 dollars from his mother and 2 more dollars from his father. How much money does he have in total?}'', the core math operation remains unchanged. However, the added narrative introduces linguistic features (e.g., sentence structure and word choice) and semantic features (e.g., familiar family roles) that may affect how examinees process the item cognitively and emotionally. While such variation can enrich individual items, when multiple items share highly similar incidental features (e.g., repeated use of ``money'' scenarios or family contexts), examinees may experience cognitive fatigue, reduced novelty, and perceive redundancy during testing (Stanton et al., 2002; Pelánek, 2019). This risk is particularly acute in domains like reading comprehension, where variety in genre, phrasing, and text type is expected (Sayin \& Gierl, 2024).

Beyond test-taking experience, incidental content similarity has demonstrable psychometric consequences. Empirical evidence suggests that high semantic similarity can compromise measurement quality. For example, Kilmen and Bulut (2025) found a significant negative correlation between the mean semantic similarity and the item discrimination parameters. Furthermore, selecting items with high semantical overlap can lead to poor model fit due to locally correlated residuals (Hommel et al., 2022), while nonsystematic item-context effects may introduce unintended error (Weinberger et al., 2006). Thus, if unaddressed, high incidental similarity can undermine the fairness and validity of ability estimates.

For CAT specifically, managing incidental content similarity is both more critical and more challenging than in traditional linear testing. In linear test forms, developers can manually curate and distribute items to ensure variety. In CAT, however, items are selected in real-time by algorithms optimized primarily for psychometric efficiency. Without explicit safeguards, a CAT algorithm is ``blind'' to incidental redundancy, it may sequentially select items that are statistically optimal but semantically or structurally repetitive, thereby amplifying the negative effects described above. To mitigate content redundancy and enhance fairness in CAT, it is essential to systematically monitor and balance incidental content similarity at both the item bank construction (Cechák \& Pelánek, 2021; Do \& Lee, 2024) and selection algorithm levels, preserving the test's discriminatory power (Kilmen \& Bulut, 2025). Such metrics also hold promise for broader applications, such as guiding content recommendations in adaptive learning systems by avoiding redundant practice and promoting diverse exposure (Cechák \& Pelánek, 2021). Consequently, achieving such managements requires a foundational capability: robust, scalable methods for quantifying incidental content similarity.

\subsection{\texorpdfstring{\emph{\textbf{2.2 Automated Item Similarity Analysis}} }{2.2 Automated Item Similarity Analysis }}\label{automated-item-similarity-analysis}

As established, effective control of incidental content redundancy in large-scale assessments require reliable metrics to quantify item similarity. Traditional approaches to item content evaluation, such as expert judgment, crowdsourcing, ablation studies, and post hoc analyses based on response data (Gorgun \& Bulut, 2024; Do \& Lee, 2024). While informative, these methods are often labor-intensive and not readily scalable for large-scale applications. For example, they offer limited support for real-time decision making within adaptive testing algorithms.

Consequently, Automated Item Similarity Analysis (AISA) methods, which assign similarity scores directly from item content, have emerged as a cost-effective foundation for item pool curation and content-based constraints in CAT. The practical value of such an approach is supported by evidence from content alignment tasks, ranking by semantic similarity yielded a 95\% probability that the correct match was among the top five candidates, drastically reducing the comparison space for experts (Butterfuss \& Doran, 2025). For CAT, an effective AISA should not only be accurate but must also produce interpretable metrics that can be translated into content constraints within selection algorithms.

Early AISA methods were largely adapted from natural language processing (NLP) and focused on surface-level lexical overlap. Metrics such as BLEU (Papineni et al., 2002) and ROUGEs (Lin, 2004), originally developed for machine translation and text summarization, quantify similarity through n-gram matching and word sequence overlap. These approaches are effective for detecting near-duplicate text and have been applied to educational items as a first step toward automated similarity assessment. However, because they primarily operate at the word or phrase level, they offer a view of similarity that is structurally narrow and semantically blind, particularly in educational contexts where paraphrasing and contextual variation are common (Gorgun \& Bulut, 2024; Moore et al., 2024).

Subsequent developments introduced greater semantic sensitivity. For instance, METEOR (Lavie \& Denkowski, 2009) extended lexical matching by incorporating stemming, synonym recognition, and paraphrase alignment, while its extension, METEOR-Next (Denkowski \& Lavie, 2010), further supported syntactic-level paraphrasing. This progression allowed for a more nuanced analysis of surface form and local phrasing, better handling lexical variation. Yet, these methods remain fundamentally rooted in aligning pre-defined lexical and syntactic units. They offer limited capacity to represent the global semantic meaning or conceptual coherence of an item as a whole, falling short of capturing the thematic or logical similarities that constitute key aspects of incidental content.

A paradigm shift arrived with the application of distributed text representations (embeddings). Metrics like cosine similarity (Rodriguez-Torrealba et al., 2022; Kilmen \& Bulut, 2025; Sayin \& Gierl, 2024; Milano et al., 2025) applied to vectors from models like word2vec or BERT (Bidirectional Encoder Representations from Transformers) marked a leap toward semantic assessment. By mapping items into a continuous semantic space, these methods can identify thematic relatedness and conceptual proximity beyond literal word overlap. However, this strength is coupled with a distinct weakness: the vector representations are opaque and often conflate stylistic features with substantive meaning. Moreover, they provide no explicit, interpretable measure of structural features like sentence complexity or item format, and their effectiveness in discerning fine-grained logical relationships within test items remains inconsistent.

\subsection{\texorpdfstring{\emph{\textbf{2.3 Large Language Models for Item Analysis}}}{2.3 Large Language Models for Item Analysis}}\label{large-language-models-for-item-analysis}

More recently, the computational text analysis field has been revolutionized by the advent of transformer-based large language models (LLMs). Trained via self-supervised learning, LLMs can generate high-dimensional text embeddings that capture intricate linguistic properties and deep contextual relationships, offering a powerful foundation for automated text analysis (Chiang \& Lee, 2023; Milano et al., 2025; Butterfuss \& Doran, 2025; Hao et al., 2024). As a result, researchers have increasingly examined LLMs as scalable complements to human judgment across educational measurement applications, including automated scoring, item generation, and the creation of ideal response references (Do \& Lee, 2024; Hao et al., 2024; Chiang \& Lee, 2023; Götz et al., 2023; Song et al., 2025; Hur \& Ji, 2025; Liu et al., 2025).

To effectively harness foundational LLMs for such domain-specific applications, strategies like prompt engineering have proven critical. By carefully designing instructional prompts, researchers can steer LLMs toward desired outputs without extensive fine-tuning, making it a resource-efficient approach for zero- or few-shot scenarios (Chen \& Wang, 2024; Hao et al., 2024). This capability is directly relevant to AISA, where crafting precise prompts can guide LLMs to evaluate text pairs in a consistent and interpretable manner.

Indeed, several recent studies have applied LLMs with prompt-based methods to assess semantic similarity in educational contexts. For instance, Hur and Ji (2025) introduced an LLM-based semantic content analysis framework aimed at stabilizing content alignment evaluation. Their method uses GPT-4o to generate ideal responses as semantic references, against which student answers are evaluated via cosine similarity. More directly focused on item comparison, Do and Lee (2024) proposed a progressive augmentation method (ProAug) for similarity labeling, using GPT-4 to analyze TOEFL-QA items. ProAug decomposes each item into three components: a detailed background narrative (averaging 3,736 characters), the question stem, and the response options. For the Question aspects pair similarity, the authors used a structured prompt: ``\emph{Please assess the similarity between question1 and question2 of TOEFL test item:\textbackslash n Question1:\{Item1 Question\}\textbackslash n Question2:\{Item2 Question\}\textbackslash n Response only with the similarity score between 0--5:}''. To provide more refined scores, they filtered pseudo-labeled outputs and validated results against expert human annotations. At the sentence level, Banjade et al. (2024) further demonstrated that LLMs can estimate semantic similarity across multiple model architectures, including GPT-3.5, GPT-4, LLAMA2, and MIXTRAL, using the given prompt ``\emph{Provide a semantic similarity score on a scale of 0 to 1, 0 being least similar and 1 being most similar, for the following two sentences \{reference sentence\} and \{student sentence\}}'' to compare reference and student sentences on a continuous similarity scale.

These works represent significant steps toward LLM-assisted item analysis, demonstrating the LLMs' ability to handle structured decomposition and similarity judgment. However, for the specific purpose of analyzing incidental content similarity, current approaches exhibit a key limitation: they treat generated components as holistic text segments rather than analytically distinguishing between the distinct structural and semantic dimensions of incidental content (e.g., scenario vs. phrasing vs. logical flow). Consequently, they do not directly support the fine-grained, multi-dimensional measurement required to effectively quantify and manage incidental redundancy in item pools.

Taken together, the progression of AISA methods points to LLMs as a flexible and scalable frontier. While existing prompt-based applications confirm LLMs utility for semantic comparisons, they also underscore the absence of explicit modeling for the multifaceted nature of incidental content. This gap motivates the present study to develop a dedicated LLM-powered framework that moves beyond holistic similarity scores toward a multidimensional and operationalizable analysis of incidental content, thereby supporting more effective content control in adaptive testing systems.

\subsection{\texorpdfstring{\emph{\textbf{2.4 Research Questions}}}{2.4 Research Questions}}\label{research-questions}

Building on the preceding sections, this study addresses incidental content redundancy as a general challenge in large-scale assessment contexts where item banks are developed, maintained, and deployed at scale. We propose an interpretable, multidimensional framework for incidental content similarity using LLMs, and evaluate its measurement and operational utility. While the framework is designed to be broadly applicable across assessment scenarios, this study focuses on CAT as a representative and methodologically demanding application context. Accordingly, this study is guided by the following three questions, which progress from conceptual framing to methodological validation and applied evaluation:

\begin{itemize}
\item
  RQ1: How can incidental content similarity be systematically conceptualized and operationalized as a multidimensional construct that distinguishes semantic and structural features of item presentation?
\item
  RQ2: To what extent can the proposed LLM-powered AISA framework generate interpretable and psychometrically meaningful similarity metrics?
\item
  RQ3: When apply to CAT, how does integrating the LLM-powered similarity metrics into a cluster-constrained item selection algorithm affect the trade-off between psychometric accuracy and content diversity during adaptive test administration?
\end{itemize}

\section{\texorpdfstring{\textbf{Methods}}{Methods}}\label{methods}

\subsection{\texorpdfstring{\emph{\textbf{3.1 Incidental Content Similarity Framework}} }{3.1 Incidental Content Similarity Framework }}\label{incidental-content-similarity-framework}

Building on prior conceptualizations of incidental content (Ford et al., 2000; Sonnleitner et al., 2025; Hur \& Ji, 2025; Gorin \& Embretson, 2006; Sayin \& Gierl, 2024) and existing approaches to item similarity analysis (Rodriguez-Torrealba et al., 2022; Kilmen \& Bulut, 2025; Do \& Lee, 2024; Lin, 2004; Denkowski \& Lavie, 2010), the present study defines incidental content as construct-irrelevant but appropriate item content that shapes item presentation and contextual framing rather than the targeted latent trait. To support systematic measurement, incidental content is operationalized as a multidimensional construct comprising two analytically distinct components:

\textbf{Structured decomposition (}\(\mathbf{S}_{\mathbf{1}}\)\textbf{).} This component captures surface-level structural similarity between items and reflects how content is organized and expressed. Analyzing the \(S_{1}\) supports the identification of repetitive presentation patterns that may increase perceived redundancy during adaptive administration. The detailed dimensions including: (1) Item format: similarity in layout or phrasing patterns (e.g., question type, use of options). (2) Sentence structure: similarity in grammatical constructions or word order. (3) Lexical overlap and paraphrasing: use of shared vocabulary, stems, or synonymous expressions.

\textbf{Semantic relatedness (}\(\mathbf{S}_{\mathbf{2}}\)\textbf{).} This component captures the underlying semantic meaning of the items and allows us to identify content overlap and redundancy at a deeper conceptual level, beyond surface-level wording. It involves three dimensions: (1) Overlapping themes or contexts: similarity in background scenarios or examples. (2) General lexical semantics: sentence-level meaning proximity between items. (3) Shared connotations or emotional tone: similarity in affective or contextual associations.

To quantify the pairwise incidental content similarity between items, a composite similarity matrix \(\mathbf{S}\) is calculated as: \(S_{ij} = w_{1}S_{1ij} + w_{2}S_{2ij}\). Where the \(S_{ij}\) is the overall similarity between item \(i\) and \(j\), \(S_{1ij}\) and \(S_{2ij}\) are the structured decomposition and semantic relatedness similarities, respectively. The \(w_{1}\) and \(w_{2}\) are domain-defined weights, with the constraint \(w_{1} + w_{2} = 1\). Different assessment scenarios may prioritize surface-level presentation diversity, deeper semantic coverage, or a balance of both. For instance, in item bank curation or AIG quality control, practitioners may prioritize semantic relatedness (\(\mathbf{S}_{\mathbf{2}}\)) to detect conceptual redundancy and thematic ``clones'', assigning it a relatively higher weight. Conversely, for assessments focused on linguistic variety, structured decomposition (\(\mathbf{S}_{\mathbf{1}}\)) might receive more emphasis. In the present study, these similarity metrics are instantiated within a CAT context, where both dimensions are considered to prevent examinee fatigue and ensure content diversity, as detailed in subsequent sections. Mathematically, since educational test items are typically related in topic or language, negative similarity values are rare (Pelánek, 2019). Therefore, all values in \(\mathbf{S}\)\textbf{,} \(\mathbf{S}_{\mathbf{1}}\)\textbf{,} \(\mathbf{S}_{\mathbf{2}}\) are non-negative.

\subsection{\texorpdfstring{\emph{\textbf{3.2 LLM-powered Automated Item Similarity Analysis Method}}}{3.2 LLM-powered Automated Item Similarity Analysis Method}}\label{llm-powered-automated-item-similarity-analysis-method}

The LLM-powered AISA method uses carefully designed prompts considering both \(\mathbf{S}_{\mathbf{1}}\) and \(\mathbf{S}_{\mathbf{2}}\) aspects, while referencing established similarity metrics (BLEU and cosine similarity) as auxiliary signals rather than primary decision rules. BLEU-based similarity was computed using the ``\emph{nltk''} Python package (Hardeniya et al., 2016), in which each item was tokenized, lowercased, and treated symmetrically as both reference and candidate across all item pairs. Cosine similarity was computed using the ``\emph{text2vec}'' R package (Welbers et al., 2017), based on a document-term matrix derived from word-tokenized item texts. Item representations were embedded in a high-dimensional lexical space and normalized using the \(L_{2}\) norm to compute the pairwise cosine similarity (Hellín et al., 2023).

In contrast to these static, token-driven measures, the LLM-powered AISA method leverages both auxiliary similarity signals and higher-order contextual interpretation. This study employs the widely adopted commercial LLM chatbots: Anthropic's Claude Sonnet 4, selected for Claude models' strong performance in structured reasoning tasks and broad accessibility in applied LLM research (Wang et al., 2025). All analyses were conducted via the model's API, with temperature fixed at 0 to ensure deterministic and reproducible outputs.

For each dimension, the LLM generates a pairwise similarity matrix (\(S_{1}\) and \(S_{2}\)) using component-specific prompts. Table 1 illustrates the prompt structure for structured decomposition. To obtain semantic relatedness scores, the component descriptions are replaced with the corresponding semantic dimensions outlined in Section 3.1, and cosine similarity is provided as the auxiliary reference instead of BLEU. Within each prompt, the LLM assigns dimension-level scores on a fixed scale (0-10) and provides a concise justification for each item pair. The resulting ranges of \(S_{ij}\), \(S_{1ij}\), and \(S_{2ij}\) are therefore 0 to 30. In this study, the weights for final \(S_{ij} = w_{1}S_{1ij} + w_{2}S_{2ij}\) were set as \(w_{1} = 0.3\) and \(w_{2} = 0.7\). These values can be adjusted depending on the intended emphasis or application.

\begin{longtable}[]{@{}
  >{\raggedleft\arraybackslash}p{(\linewidth - 2\tabcolsep) * \real{0.1538}}
  >{\raggedright\arraybackslash}p{(\linewidth - 2\tabcolsep) * \real{0.8462}}@{}}
\caption{Prompt Template for Structured Decomposition Similarity Calculation}\tabularnewline
\toprule\noalign{}
\begin{minipage}[b]{\linewidth}\raggedleft
Role Setup:
\end{minipage} & \begin{minipage}[b]{\linewidth}\raggedright
You are a researcher working on one study of content analysis of survey. You aim to evaluate the similarity of the item content.
\end{minipage} \\
\midrule\noalign{}
\endfirsthead
\toprule\noalign{}
\begin{minipage}[b]{\linewidth}\raggedleft
Role Setup:
\end{minipage} & \begin{minipage}[b]{\linewidth}\raggedright
You are a researcher working on one study of content analysis of survey. You aim to evaluate the similarity of the item content.
\end{minipage} \\
\midrule\noalign{}
\endhead
\bottomrule\noalign{}
\endlastfoot
Task: & Using the 36-items the Experiences in Close Relationships (ECR) scale shown in \textless item samples\textgreater. Calculate pairwise \textbf{Structured decomposition} similarity between items based on \textless rules\textgreater. \\
Components: & \begin{minipage}[t]{\linewidth}\raggedright
\begin{enumerate}
\def\labelenumi{\arabic{enumi}.}
\item
  \emph{Item format (10 points)}: similarity in layout or phrasing patterns (e.g., question type, use of options).
\item
  \emph{Sentence structure (10 points)}: similarity in grammatical constructions or word order.
\item
  \emph{Lexical overlap and paraphrasing (10 points)}: use of shared vocabulary, stems, or synonymous expressions.
\end{enumerate}
\end{minipage} \\
Rules: & \begin{minipage}[t]{\linewidth}\raggedright
\begin{itemize}
\item
  Take the provided \textless BLEU similarity\textgreater{} matrix as an extra information into consideration, and do not recalculate the BLEU.
\item
  The final \textbf{Structured decomposition} similarity relies on your own judgment on the three \textless Components\textgreater{} in this prompt.
\item
  The final similarity is number ranging from 0 to 30, which adds up all scores of components together. For example, if two items have 4 points, 3 points, 3 points for three components. The final similarity score of structured decomposition is 10 points. Lower value indicating lower similarity of structured decomposition and semantic relatedness or vice versa.
\end{itemize}
\end{minipage} \\
Output: & Return a JSON array of objects, each containing: `item\_i', `item\_j', `score on Overlapping themes or contexts', `score on General lexical semantics', `score on Shared connotations or emotional tone', `final\_similarity', `explanation' (a brief but important explanation of how the scores were determined). \\
\end{longtable}

\subsection{\texorpdfstring{\emph{\textbf{3.3 Data Samples and Measurement Model}}}{3.3 Data Samples and Measurement Model}}\label{data-samples-and-measurement-model}

This study used items from the short form of the Experiences in Close Relationships (ECR) scale as an empirical testbed for the proposed item similarity analysis framework. The ECR provides a well-established, theory-driven measure with rich item-level semantic variation, making it particularly suitable for evaluating automated similarity methods that aim to capture both structural and semantic aspects of incidental content. The scale consists of 36 five-point Likert-type items that assess attachment-related anxiety and avoidance in romantic relationships (Kilmen \& Bulut, 2025). Item details are provided in the Appendix. The response data are available on the Open-Source Psychometric Project website (\url{https://openpsychometrics.org/_rawdata/}). The raw dataset includes 51,491 participants across various countries, including United States, Palau, Macao, Japan, Italy, China and more.

Before formal modeling, the data were cleaned to remove participants outside the 18-65 age range and those with missing responses for any of the 36 items, remaining 35,278 valid participants. Response values of 0 were treated as missing, and items that were reverse scored according to ECR conventions (Q3, Q15, Q19, Q22, Q25, Q27, Q29, Q31, Q33) were recoded so that higher scores consistently reflected higher levels of the latent traits.

The measurement model was the graded response model (GRM), suitable for Likert-type items, and was fitted using the ``\emph{mirt}'' package in R (Chalmers, 2012). To ensure model stability and prevent implausible estimates, constraints were applied to the discrimination parameters (\(\alpha \geq\) 0.1) and threshold parameters (\(b \in \lbrack - 10,10\rbrack\)). Model estimation was conducted using the Metropolis-Hastings Robbins-Monro (MHRM) algorithm, and participants' latent trait scores were obtained via expected a posteriori (EAP) estimation.

\subsection{\texorpdfstring{\emph{\textbf{3.4 Validation of LLM-Powered Item Similarity Metrics}}}{3.4 Validation of LLM-Powered Item Similarity Metrics}}\label{validation-of-llm-powered-item-similarity-metrics}

To evaluate the interpretability and reliability of the LLM-powered AISA method (RQ2), this study conducted validation by comparing the LLM-generated similarity with BLEU and cosine similarity from a psychometric perspective.

\emph{\textbf{Correspondence with item residual correlations.}} The residual correlations extracted from GRM model and captured construct-irrelevant local dependencies among items, which may arise when items share overlapping incidental content (Christensen et al., 2017). Spearman rank correlations were calculated under the assumption that higher incidental content similarity scores are associated with stronger local dependencies, such validation focused on monotonic correspondence between similarity rankings rather than linear associations between raw metric values.

\emph{\textbf{Key item characteristics within similarity-based clusters.}} To further validate the LLM-generated similarity, items were grouped into clusters based on their similarity scores using hierarchical clustering, which is designed to aggregate items with higher mutual similarity while separating dissimilar items into different clusters (Murtagh \& Contreras, 2012). We then examined whether items within the same cluster displayed coherent psychometric properties, including item discrimination and GRM thresholds.

\subsection{\texorpdfstring{\emph{\textbf{3.5 Similarity-Constrained CAT Selection Algorithm}} }{3.5 Similarity-Constrained CAT Selection Algorithm }}\label{similarity-constrained-cat-selection-algorithm}

Although the proposed item similarity framework is designed to support multiple assessment scenarios, this section focuses on its integration into CAT to illustrate its operational impact in a dynamic item selection environment. To integrate the LLM-powered similarity metrics into a CAT selection algorithm (RQ3), this study used similarity-based clustering as a practical constraint. Prior research has shown that item clustering can support the definition or refinement of knowledge components and has been widely adopted in recommender systems (Rihák \& Pelánek, 2017). Building on this rationale, similarity clusters were treated as operational constraints within CAT. The constrained selection procedure was implemented using the two-phase item selection strategy (Cheng et al., 2007), combined with the maximum priority index (MPI) method (Cheng \& Chang, 2009), which allows flexible control of lower and upper bounds for each cluster.

To enable clustering, the LLM-generated similarity matrix \(\mathbf{S}\) was transformed into a distance matrix \(\mathbf{D}\) using \(D_{ij} = 1 - \frac{S_{ij} - Min(S_{ij})}{Max\left( S_{ij} \right) - Min(S_{ij})}\), where higher similarity results in smaller distances. This transformation supports the application of clustering, which can enhance retrieval efficiency by reducing the search space for subsequent item selection (Chen \& Wang, 2024).

To evaluate the effectiveness of the proposed cluster-constrained CAT algorithm, a simulation study was conducted with 50 replications, each involving 2,000 examinees and a fixed test length of 20 items. The traditional Maximum Fisher Information (\textbf{MFI}) method, a widely used criterion in CAT for maximizing information at the examinee's current ability estimate (Chang, 2015), was used as the baseline in this study. The key manipulation was the type of similarity metric used to form clusters (LLM-generated, BLEU, Cosine), each fed into the same similarity-constrained MFI selection algorithm (SC-MFI). Under the SC-MFI condition, a lower bound of 3 items per cluster was strictly enforced to reduce the risk of presenting incidental content redundant items, while an upper bound of 10 items per cluster allowed sufficient flexibility in adaptive selection. The performance of each selection strategy was evaluated using the root mean square error (RMSE) and bias of the ability estimates.

\section{\texorpdfstring{\textbf{Results}}{Results}}\label{results}

\subsection{\texorpdfstring{\emph{\textbf{4.1 Decomposing Incidental Content Similarity: Semantic vs. Structural Dimensions}}}{4.1 Decomposing Incidental Content Similarity: Semantic vs. Structural Dimensions}}\label{decomposing-incidental-content-similarity-semantic-vs.-structural-dimensions}

Based on the structured prompt design, the LLM generated similarity scores separately for structured decomposition (\(S_{1}\)) and semantic relatedness (\(S_{2}\)). The total LLM-generated \(\mathbf{S}\) scores ranged from 6 to 26.60 (\(M = 15.41,SD = 4.02\)), which exhibited a high correlation with \(S_{2}\) (\(r = \ \)0.98), moderately with \(S_{1}\) (\(r = \ \)0.76), consistent with the weight settings used in this study (\(w_{1} = 0.3,w_{2} = 0.7\) ). Additionally, the total \(\mathbf{S}\) scores were less strongly correlated with traditional text-based metrics BLEU (\(r = \ \)0.30) and cosine similarity (\(r = \ \)0.43). These small to moderate correlations indicates partial overlap with traditional text-based metrics, while also suggesting that the LLM captures additional structural and semantic aspects not fully reflected by BLEU and cosine similarity.

Across all item pairs, \(S_{1}\) scores ranged from 6 to 28 (\(M = 15.03,SD = 3.28\)), while \(S_{2}\) scores ranged from 6 to 27 (\(M = 15.57,SD = 4.66\)). Within \(S_{1}\), item format similarity had the highest average score (\(M = 7.97,SD = 0.54\)), followed by sentence structure (\(M = 4.13,SD = 1.62\)) and lexical overlap or paraphrasing (\(M = 2.93,SD = 1.69\)). This pattern reflects a common feature of operational assessments and is consistent with item modeling approaches for automatic item generation (Fu et al., 2022; Embretson \& Kingston, 2018), in which items frequently reuse presentation templates and syntactic structures while minimizing direct lexical repetition. Within \(S_{2}\), overlapping themes or contexts showed the highest average similarity (\(M = 6.81,SD = 1.22\)), followed by shared connotations or emotional tone (\(M = 4.98,SD = 2.66\)), and general lexical semantics (\(M = 3.79,SD = 1.71\)). This distribution suggests that semantic similarity within an assessment may primarily drive by shared scenarios rather than direct sentence-level semantic equivalence, which is consistent with intentional test design practices (Milano et al., 2025).

The Spearman rank correlation between \(S_{1}\) and \(S_{2}\) was 0.66, indicating general alignment while preserving meaningful differentiation between surface-level structure and deeper semantic content. Figure 1 visualizes the pairwise discrepancies between \(S_{1}\) and \(S_{2}\). Most item pairs fall within a {[}-5,5{]} difference range, suggesting that, for many item pairs, structural and semantic similarity evolve in parallel. Importantly, a subset of item pairs exhibits larger discrepancies. To assess the interpretability of such discrepancies, these item pairs were examined based on the LLM's chain-of-thought rationales and the detailed outputs.

\includegraphics[width=5.39345in,height=2.85in,alt={A diagram of a graph AI-generated content may be incorrect.}]{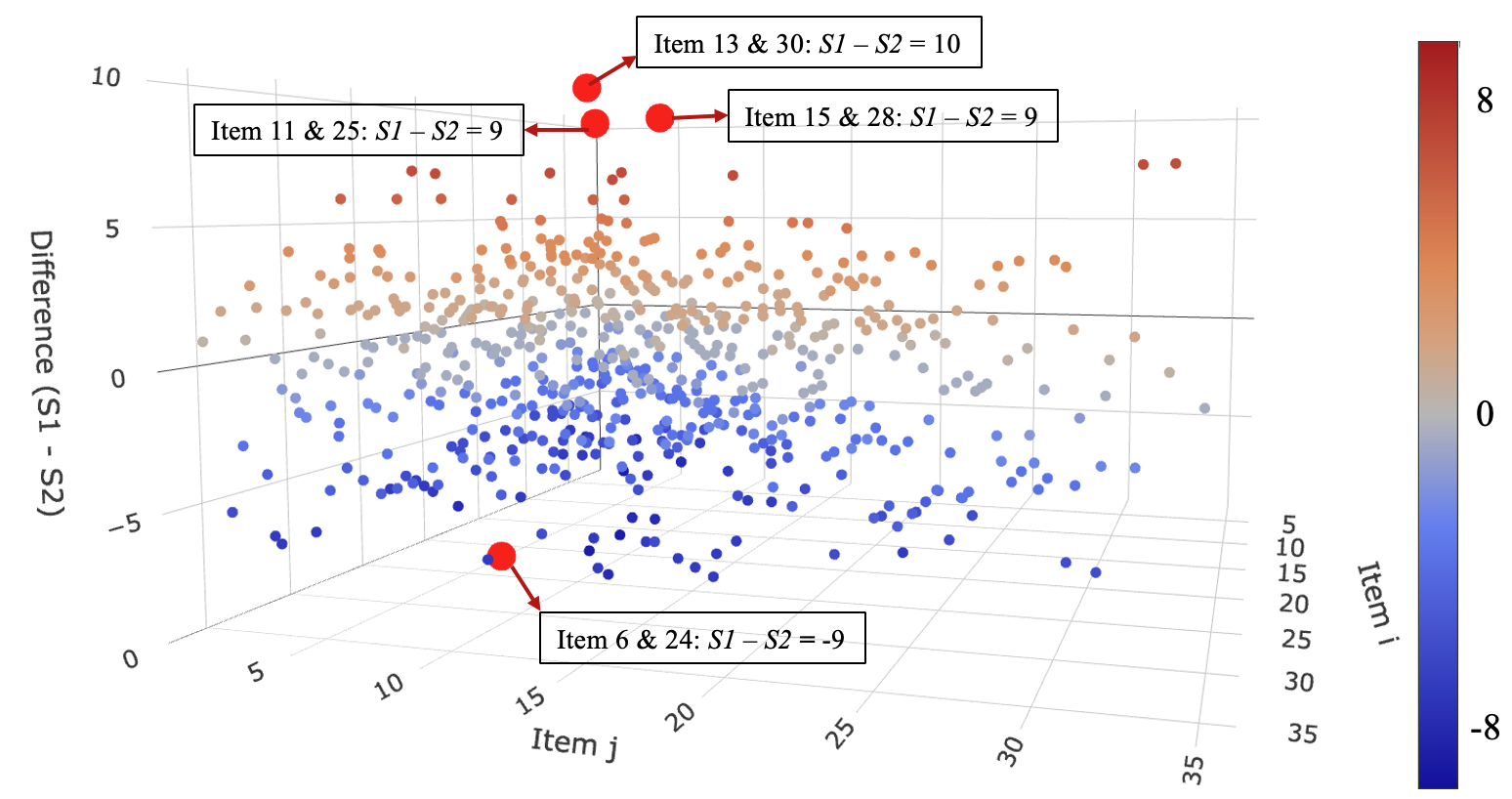}

Figure 1. Difference Between Similarity Scores for Two Dimensions across item pairs.

For example, the pair item 13 (``\emph{I am nervous when partners get too close to me}'') and item 30 (``\emph{I get frustrated when my partner is not around as much as I would like}'') demonstrated the largest difference: \(S_{1} = 16\ \)and \(S_{2} = 6\), the details are illustrated in Figure 2. As shown in the radar chart, for the relatively high structured decomposition (\(S_{1}\)), Claude Sonnet 4 highlighted that both items share a similar declarative statement format with emotional responses to relationship situations (8/10 for format), employ similar sentence structure with ``I {[}emotion verb{]} when {[}partner situation{]}'' pattern (6/10 for structure), but have minimal lexical overlap, sharing only basic words such as ``I'', ``when'', and ``partner/partners'' (2/10 for lexical overlap). For semantic relatedness (\(S_{2}\)), Claude Sonnet 4 indicated that although both items address relationship proximity concerns but from opposite perspectives, item 13 reflects avoidance anxiety about too much closeness while item 30 reflects anxious attachment about insufficient closeness, they have different emotional responses (nervousness VS. frustration) and opposite desired outcomes.

\includegraphics[width=4.24574in,height=2.63889in,alt={A diagram of a diagram AI-generated content may be incorrect.}]{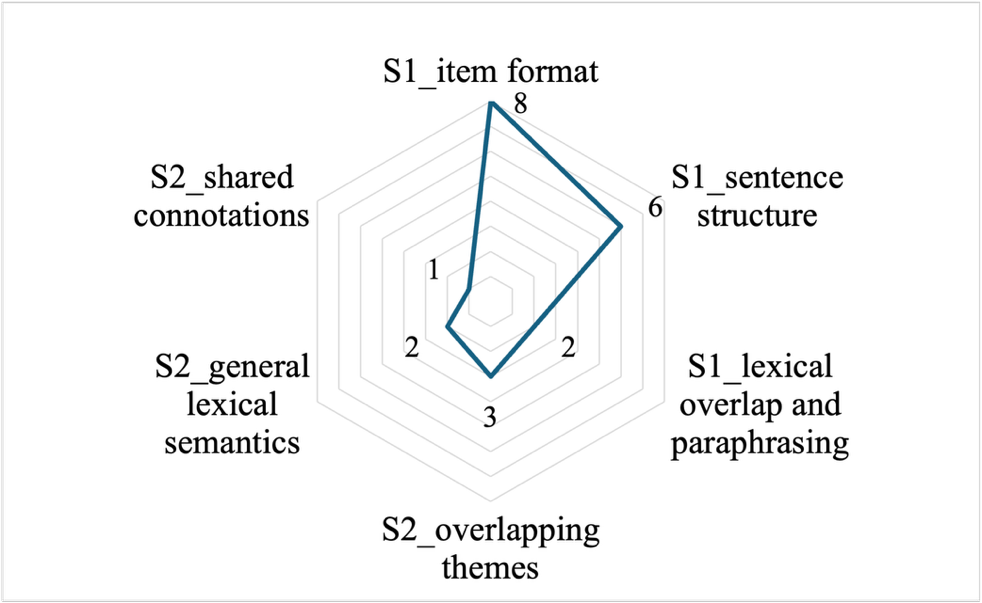}

Figure 2. Dual-dimensional Similarity Profile for Sample Item Pair (Items 13 and 30).

\subsection{\texorpdfstring{\emph{\textbf{4.2 Performance of the LLM-Powered AISA Similarity Metrics}} }{4.2 Performance of the LLM-Powered AISA Similarity Metrics }}\label{performance-of-the-llm-powered-aisa-similarity-metrics}

To evaluate the performance of the LLM (Claude Sonnet 4) from a psychometric perspective, a GRM was first fitted to the valid ECR item responses. The model was converged within a 0.001 tolerance after 11 MHRM iterations, with diagnostic checks confirming stable parameter estimation. The resulting latent trait approximated a standard normal distribution (\(M = 0.01,SD = 0.96\)), while the item discrimination parameters ranged broadly, reflecting differential informativeness across the items. More information is shown in Appendix (Table II).

The residual correlation matrix (Christensen et al., 2017) was extracted, Spearman correlations were computed between the residual correlation values and three similarity metrics: BLEU, cosine similarity, and the LLM-generated total similarity scores. Results showed weak associations for traditional metrics (BLEU: \(r = - 0.10\); cosine: \(r = 0.06\)), while the LLM-generated similarity exhibited a higher correlation (\(r = 0.50\)). An \(r = 0.50\) is not ``only moderate'' in this context, since residual correlations are noisy indicators of incidental content similarity because they also capture, such as testlet effects and sampling variability. So the theoretical ceiling for this correlation is well below 1. An \(r = 0.50\) against a noisy criterion, compared to essentially zero for the alternatives, is strong evidence of criterion-related validity. This finding indicates that the LLM-powered AISA captures patterns of incidental content similarity that align more closely with psychometrically meaningful local item dependencies than BLEU and cosine similarity.

Based on the final similarity matrices, hierarchical clustering was performed to identify item groupings, such that items with higher pairwise similarity were preferentially grouped together, while clusters were separated by relatively lower similarity. Figure 3 presents the resulting dendrogram based on LLM-generated final similarity, BLEU and cosine results are attached in the appendix (Figure I and II).

In this study, the 36 ECR items were partitioned into 4 clusters to achieve meaningful similarity differentiation while maintaining enough items per cluster for subsequent CAT implementation. For the LLM-based clustering, items within the same cluster showed higher internal similarity than the overall item pool, with mean within-cluster similarity ranging from 19.2 to 22.0, compared to a global mean of 15.41 (\(SD = 4.02\)). Additionally, the clustering results indicated that BLEU- and cosine-based approaches tended to assign a large proportion of items to a single dominant cluster, whereas the LLM-based approach produced a more balanced and differentiated partition of the item pool.

\includegraphics[width=4.29167in,height=2.2004in,alt={A diagram of a diagram AI-generated content may be incorrect.}]{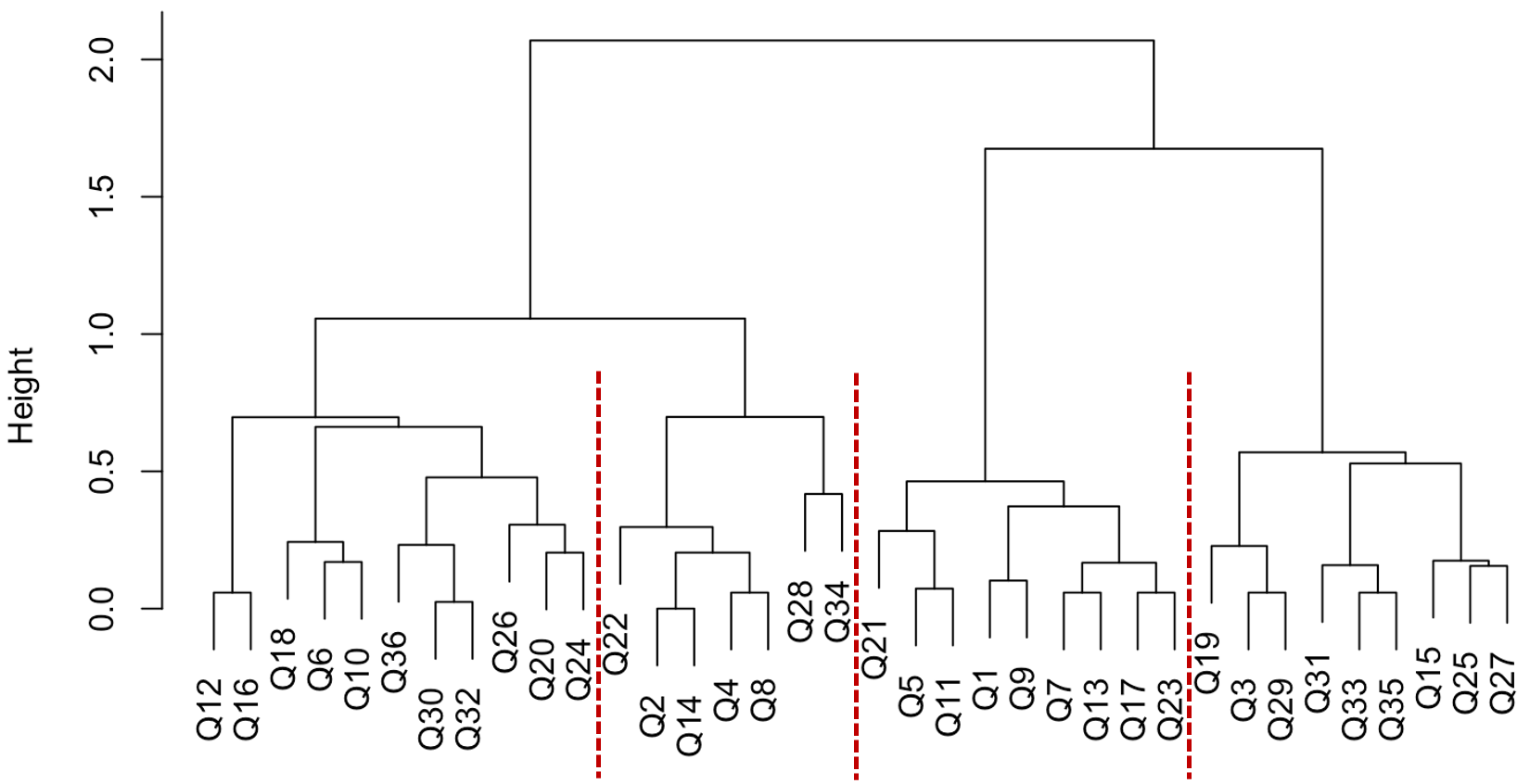}

Figure 3. Hierarchical cluster results for final LLM-generated similarity.

As shown in Figure 4, the incidental content similarity exhibits a systematic association with item discrimination, which is broadly consistent with prior findings that semantic similarity can be related to item functioning (Kilmen \& Bulut, 2025). Specifically, although BLEU- and cosine-based clustering each produced one cluster dominated by low-discrimination items, the overall discrimination structure across clusters was less clearly differentiated. In contrast, the clusters derived from LLM-generated similarity demonstrated a more interpretable and coherent pattern: aside from one outlier (Item 35), which exhibited a highly skewed response distribution and non-informative discrimination, two clusters predominantly consisted of items with very low discrimination parameters, whereas the remaining two clusters were primarily composed of relatively highly discriminating items. Items within the same LLM cluster behave similarly in terms of how well they differentiate examinees.

\includegraphics[width=4.57955in,height=2.91408in,alt={A diagram of a cluster AI-generated content may be incorrect.}]{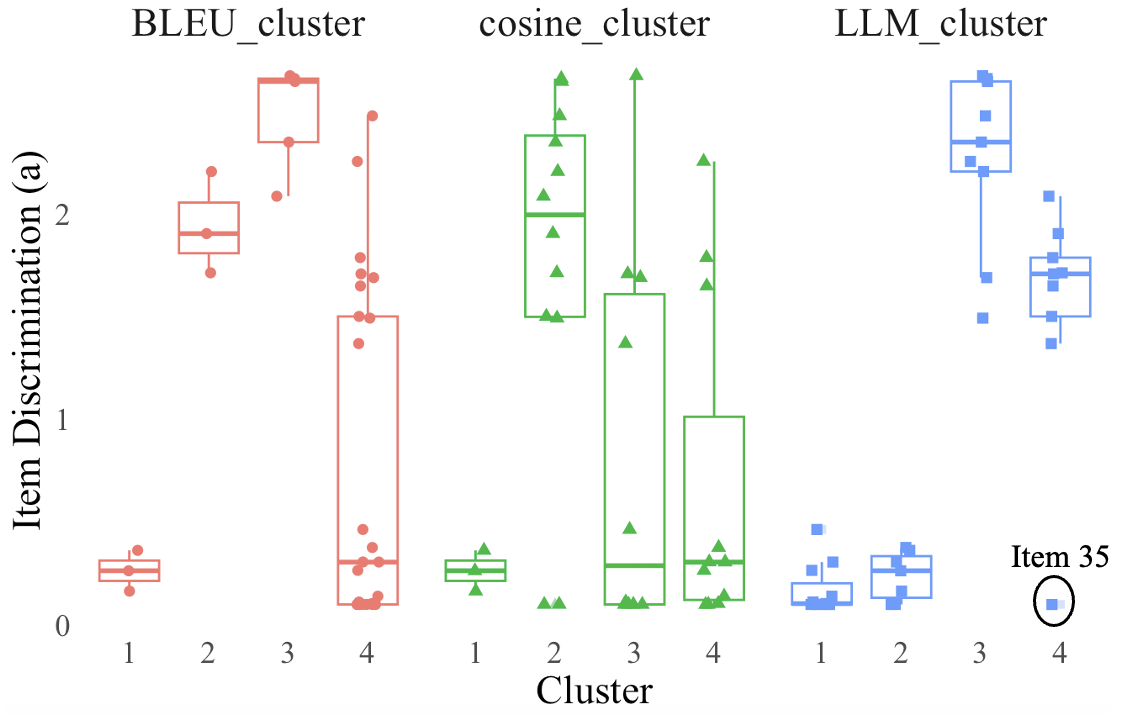}

Figure 4. Item Discrimination Distribution by Different Similarity Generation Methods.

Furthermore, a MANOVA was performed on the four GRM threshold parameters (\(b_{1} - b_{4}\)), with cluster membership from each method serving as the independent factor. The analysis revealed a stronger overall effect for the clusters derived from LLM-based similarity (\emph{F}(12, 93) = 4.79, \emph{P} \textless{} .001) compared with BLEU-based (\emph{P} = .042) and cosine-based clusters (\emph{P} = .017). Follow-up univariate ANOVAs clarified that the LLM-based clusters differed significantly on the first (\(b_{1}\)) and last (\(b_{4}\)) thresholds, whereas the intermediate thresholds (\(b_{2}\), \(b_{3}\)) showed no significant differences. These results suggested that the LLM-driven clustering effectively groups items based on features that systematically influence the endorsement of the most extreme response categories.

\subsection{\texorpdfstring{\emph{\textbf{4.3 Impact of Similarity Constraints on CAT Performance}}}{4.3 Impact of Similarity Constraints on CAT Performance}}\label{impact-of-similarity-constraints-on-cat-performance}

To reduce the potential negative effects of incidental content repetition, similarity-based constraints were incorporated into the CAT item selection procedure. Scatterplots of 50 replications bias and RMSE for the different selection strategies are presented in Figure 5.

The LLM-based SC-MFI condition achieved the lowest average bias (\(M = 0.029\)), indicating minimal systematic distortion in ability estimation. BLEU- (\(M = 0.033\)) and cosine-constrained (\(M = 0.031\)) clusters produced slightly higher bias, though still comparable to or smaller than MFI (\(M = 0.033\)). Regarding RMSE, the unconstrained MFI showed the lowest mean (\(M = 0.106\)), whereas the LLM-constrained RMSE was higher (\(M = 0.115\)) but exhibited the most concentrated distribution, reflecting greater stability across replications. BLEU- (\(M = 0.113\)) and cosine-based (\(M = 0.112\)) constraints displayed intermediate RMSE levels with wider dispersion, less consistent performance compared to LLM-constrained RMSE.

\includegraphics[width=4.70734in,height=3.04167in,alt={A group of black dots with red dots AI-generated content may be incorrect.}]{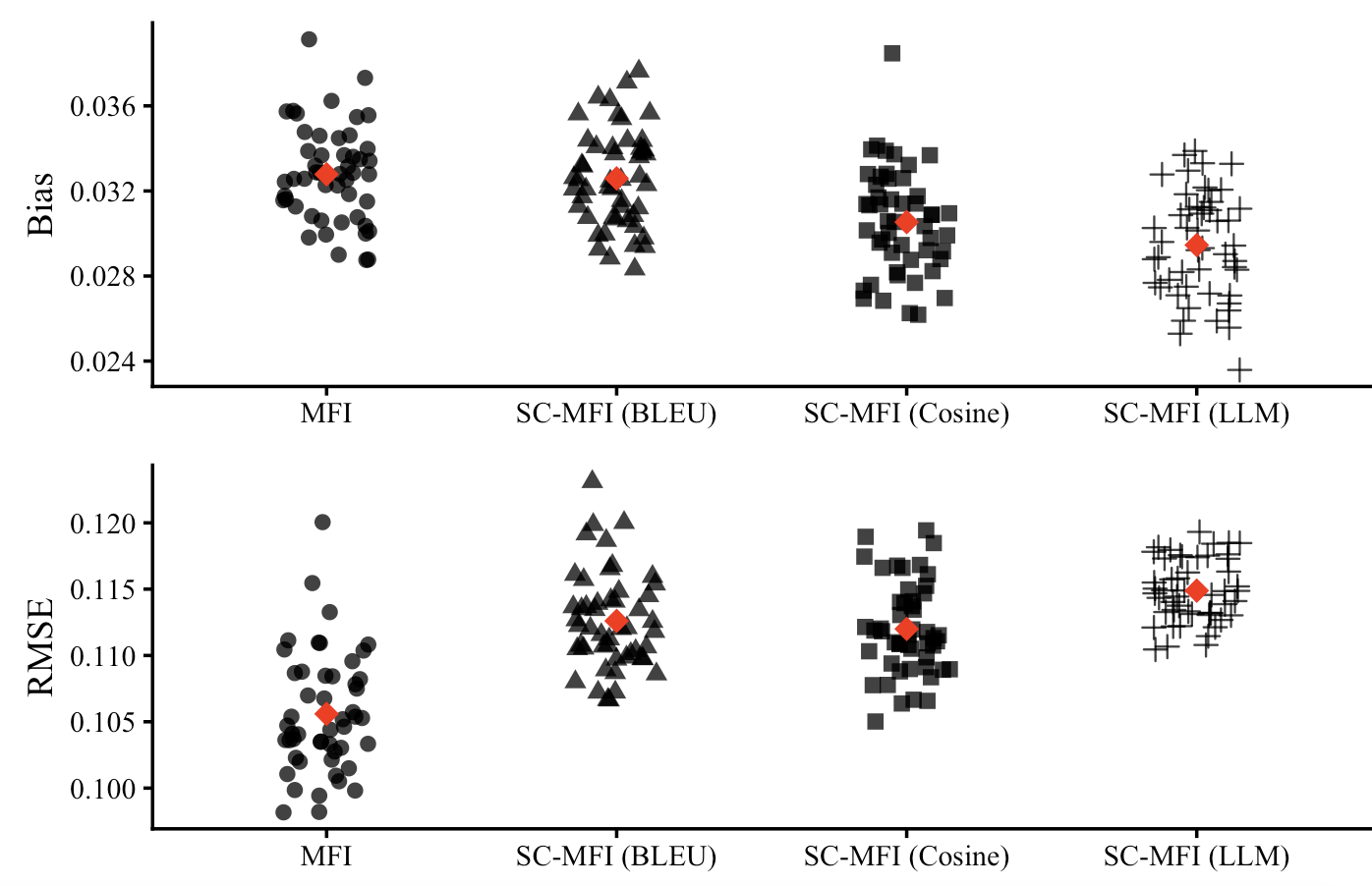}

Figure 5. Comparison of Bias and RMSE Across Item Selection Methods.

\section{\texorpdfstring{\textbf{Discussion and Conclusions}}{Discussion and Conclusions}}\label{discussion-and-conclusions}

This study advances prior work on incidental content by proposing and empirically operationalizing a dual-dimensional similarity framework, distinguishing surface-level structural similarity from deeper semantic relatedness. Previous studies have acknowledged the role of incidental content in inducing local item dependence and response fatigue (Ford et al., 2000; Sonnleitner et al., 2025; Hur \& Ji, 2025; Gorin \& Embretson, 2006; Sayin \& Gierl, 2024), yet existing automatic similarity approaches have largely relied on surface textual overlap or distributional semantics alone (Rodriguez-Torrealba et al., 2022; Kilmen \& Bulut, 2025; Do \& Lee, 2024; Lin, 2004; Denkowski \& Lavie, 2010). Such methods may capture fragments of incidental content but do not fully reflect its multifaceted nature.

\subsection{\texorpdfstring{\emph{\textbf{5.1 Theoretical Contributions: Advancing the Conceptualization of Incidental Content}}}{5.1 Theoretical Contributions: Advancing the Conceptualization of Incidental Content}}\label{theoretical-contributions-advancing-the-conceptualization-of-incidental-content}

By leveraging a structured prompt design, the LLM-based AISA method was able to generate interpretable pairwise similarity scores that reflect both dimensions simultaneously. The moderate correlation between structural and semantic similarity (\(r = 0.66\)) suggests meaningful alignment while preserving substantive differentiation, supporting the theoretical distinction between these two components. Moreover, the relatively weak associations between the total LLM-generated similarity and traditional metrics (BLEU: \(r = \ \)0.30 and cosine: \(r = \ \)0.43) indicate that the proposed approach captures additional information beyond conventional text-based similarity, even though those metrics were explicitly incorporated into the prompt. These findings suggest that LLMs can function as integrative similarity analyzers rather than simple proxies for existing metrics.

Importantly, the stronger correspondence between LLM-generated similarity and residual correlations, compared to BLEU and cosine, provides psychometric evidence that the proposed similarity scores align more closely with construct-irrelevant local dependencies. Beyond pairwise similarity, this study demonstrated that clusters derived from LLM-generated similarity exhibited coherent psychometric structure, particularly with respect to item discrimination and threshold parameters. While prior work has reported associations between semantic similarity and item functioning (e.g., Kilmen \& Bulut, 2025), the present findings extend this literature by showing that LLM-based clustering yields clearer and more interpretable differentiation among items than clustering based on BLEU or cosine similarity.

\subsection{\texorpdfstring{\emph{\textbf{5.2 Practical Implications: A Versatile Toolkit for Assessment Quality}}}{5.2 Practical Implications: A Versatile Toolkit for Assessment Quality}}\label{practical-implications-a-versatile-toolkit-for-assessment-quality}

By formalizing incidental content as a dual-dimensional construct, this study offers a practical toolkit that extends beyond any single assessment format. As contemporary item banks grow in scale and are increasingly populated through automated or AI-assisted processes, under such conditions, manual review of incidental content becomes infeasible, and traditional categorical content controls are insufficient to capture subtle but consequential forms of redundancy. The proposed LLM-powered AISA framework addresses this gap by enabling scalable, fine-grained monitoring of incidental content similarity across diverse assessment contexts.

First, the framework serves as an efficient diagnostic tool for item bank curation and quality control. As AIG continues to accelerate the production of new content (Götz et al., 2023; Song et al., 2025), the risk of ``item cloning'', where new items are merely superficial variations of existing ones, becomes a significant threat. By applying the \(S_{1}\) and \(S_{2}\) metrics during the pre-testing phase, developers can automatically identify and filter out redundant items that lack sufficient novelty. This ensures that large-scale banks remain diverse and robust, reducing the burden on human expert reviewers (Cechák \& Pelánek, 2021; Do \& Lee, 2024).

Second, the framework enables ``content-aware'' test assembly and administration. In settings such as ranking tasks or forced-choice formats (Zheng et al., 2024), subtle semantic overlaps can inadvertently guide response behavior or increase cognitive load. The LLM-powered AISA provides a principled mechanism to monitor and control these overlaps, ensuring that the presentation of content remains engaging and free from distracting repetitions.

Finally, the operational value of this framework is most clearly demonstrated through its implementation in CAT. Our simulation results show that although unconstrained MFI produced the lowest average RMSE, the LLM-based SC-MFI condition achieved the lowest bias and the most stable RMSE distribution across replications. This pattern suggests that similarity constraints introduce a modest efficiency trade-off but improve robustness and consistency, which are often desirable properties in operational testing contexts. In contrast, BLEU- and cosine-based constraints showed less stable performance, likely reflecting weaker alignment between their similarity structures and underlying psychometric properties. These findings imply that incorporating LLM-based similarity constraints may be particularly valuable in contexts where test experience quality, content diversity, and fairness are prioritized alongside estimation accuracy. Rather than optimizing information at the expense of content redundancy, similarity-constrained CAT offers a principled mechanism to manage incidental content exposure.

\subsection{\texorpdfstring{\emph{\textbf{5.3 Limitations and Future Directions}}}{5.3 Limitations and Future Directions}}\label{limitations-and-future-directions}

While this study provides robust evidence for the utility and validity of the LLM-powered AISA framework, several limitations warrant acknowledgment and suggest avenues for future research. First, this study employed a single LLM (Claude Sonnet 4) and a single item bank (the ECR scale) for validation. Different LLMs may exhibit varying strengths in semantic understanding, structural analysis, or reasoning consistency. Future research should conduct cross-model comparisons, evaluating the performance of alternative LLMs (e.g., GPT-4, Gemini, LLaMA-based models) on the same or diverse item pools to assess the stability and robustness of similarity judgments across architectures. Second, while the prompt design emphasized interpretability and the weights were set based on theoretical considerations, these choices were somewhat arbitrary and may not be optimal for all contexts and domains. Thus, alternative weighting schemes or prompt structures may further refine the balance between structural and semantic similarity.

\section*{References}

Al-Shamri, M. Y. H. (2014). Power coefficient as a similarity measure for memory-based collaborative recommender systems. \emph{Expert Systems with Applications, 41}(13), 5680-5688.

Banjade, R., Oli, P., Sajib, M. I., \& Rus, V. (2024, July). Identifying gaps in students' explanations of code using llms. In \emph{International Conference on Artificial Intelligence in Education} (pp. 268-275). Cham: Springer Nature Switzerland.

Beavers, A. S., Lounsbury, J. W., Richards, J. K., Huck, S. W., Skolits, G. J., \& Esquivel, S. L. (2013). Practical considerations for using exploratory factor analysis in educational research. \emph{Practical assessment, research \& evaluation, 18}(6), n6.

Butterfuss, R., \& Doran, H. (2025). An application of text embeddings to support alignment of educational content standards. \emph{Educational Measurement: Issues and Practice, 44}(1), 73-83.

Cechák, J., \& Pelánek, R. (2021). Experimental Evaluation of Similarity Measures for Educational Items. \emph{International Educational Data Mining Society}.

Chalmers, R. P. (2012). mirt: A multidimensional item response theory package for the R environment. \emph{Journal of statistical Software, 48}, 1-29.

Chang, H-H. (2015). Psychometrics behind computerized adaptive testing. \emph{Psychometrika, 80}(1), 1-20.

Chen, D., \& Wang, J. (2024). A Prompt Example Construction Method Based on Clustering and Semantic Similarity. \emph{Systems, 12}(10), 410.

Cheng, Y., \& Chang, H-H. (2009). The maximum priority index method for severely constrained item selection in computerized adaptive testing. \emph{British journal of mathematical and statistical psychology, 62}(2), 369-383.

Cheng, Y., Chang, H-H., \& Yi, Q. (2007). Two-phase item selection procedure for flexible content balancing in CAT. \emph{Applied Psychological Measurement, 31}(6), 467-482.

Chiang, C. H., \& Lee, H. Y. (2023). Can large language models be an alternative to human evaluations?. \emph{arXiv preprint arXiv:2305.01937}.

Christensen, K. B., Makransky, G., \& Horton, M. (2017). Critical values for Yen's Q3: Identification of local dependence in the Rasch model using residual correlations. \emph{Applied psychological measurement, 41}(3), 178-194.

Denkowski, M., \& Lavie, A. (2010, June). Extending the METEOR machine translation evaluation metric to the phrase level. In \emph{Human Language Technologies: The 2010 Annual Conference of the North American Chapter of the Association for Computational Linguistics} (pp. 250-253).

Ding, C. S., \& Hershberger, S. L. (2002). Assessing content validity and content equivalence using structural equation modeling. \emph{Structural Equation Modeling, 9}(2), 283-297.

Do, H., \& Lee, G. G. (2024, July). Aspect-Based Semantic Textual Similarity for Educational Test Items. In \emph{International Conference on Artificial Intelligence in Education} (pp. 344-352). Cham: Springer Nature Switzerland.

Embretson, S. E., \& Kingston, N. M. (2018). Automatic item generation: A more efficient process for developing mathematics achievement items?. \emph{Journal of Educational Measurement, 55}(1), 112-131.

Ford, J. M., Stetz, T. A., Bott, M. M., \& O'Leary, B. S. (2000). Automated content analysis of multiple-choice test item banks. \emph{Social science computer review, 18}(3), 258-271.

Fu, Y., Choe, E. M., Lim, H., \& Choi, J. (2022). An evaluation of automatic item generation: A case study of weak theory approach. \emph{Educational Measurement: Issues and Practice, 41}(4), 10-22.

Gierl, M. J., \& Lai, H. (2016). A process for reviewing and evaluating generated test items. \emph{Educational Measurement: Issues and Practice, 35}(4), 6-20.

Gorgun, G., \& Bulut, O. (2024). Current evaluation methods are a bottleneck in automatic question generation. \emph{In AI for education: Bridging innovation and responsibility at the 38th AAAI annual conference on AI}.

Gorin, J. S., \& Embretson, S. E. (2006). Item difficulty modeling of paragraph comprehension items. \emph{Applied Psychological Measurement, 30}(5), 394-411.

Götz, F. M., Maertens, R., Loomba, S., \& van der Linden, S. (2023). Let the algorithm speak: How to use neural networks for automatic item generation in psychological scale development. \emph{Psychological Methods, 29}(3), 494--518.

Hao, J., von Davier, A. A., Yaneva, V., Lottridge, S., von Davier, M., \& Harris, D. J. (2024). Transforming assessment: The impacts and implications of large language models and generative AI. \emph{Educational Measurement: Issues and Practice, 43}(2), 16-29.

Hardeniya, N., Perkins, J., Chopra, D., Joshi, N., \& Mathur, I. (2016). \emph{Natural language processing: python and NLTK.} Packt Publishing Ltd.

Haroz, E. E., Kane, J. C., Nguyen, A. J., Bass, J. K., Murray, L. K., \& Bolton, P. (2020). When less is more: reducing redundancy in mental health and psychosocial instruments using Item Response Theory. \emph{Global Mental Health, 7}, e3.

Hellín, C. J., Valledor, A., Cuadrado-Gallego, J. J., Tayebi, A., \& Gómez, J. (2023). A Comparative Study on R Packages for Text Mining. \emph{IEEE Access, 11}, 99083-99100.

Hommel, B. E., Wollang, F. J. M., Kotova, V., Zacher, H., \& Schmukle, S. C. (2022). Transformer-based deep neural language modeling for construct-specific automatic item generation. \emph{psychometrika, 87}(2), 749-772.

Huang, J., Zhang, Y., Morphew, J. W., Nissen, J. M., Van Dusen, B., \& Chang, H-H. (2025). Two‐Phase Content‐Balancing CD‐CAT Online Item Calibration. \emph{Journal of Educational Measurement. 62}(4), 787-808.

Hur, W., \& Ji, B. (2025). A Hybrid System for Automated Assessment of Korean L2 Writing: Integrating Linguistic Features with LLM. \emph{Systems, 13}(10), 851.

Kilmen, S., \& Bulut, O. (2025). Shortening Psychological Scales: Semantic Similarity Matters. \emph{Educational and Psychological Measurement}, 00131644251319047.

Lavie, A., \& Denkowski, M. J. (2009). The METEOR metric for automatic evaluation of machine translation. \emph{Machine translation, 23}, 105-115.

Li, H., Dong, Q., Chen, J., Su, H., Zhou, Y., Ai, Q., Ye, Z., \& Liu, Y. (2024). Llms-as-judges: a comprehensive survey on llm-based evaluation methods\emph{. arXiv preprint arXiv:2412.05579}.

Lin, C. Y. (2004, July). Rouge: A package for automatic evaluation of summaries. In \emph{Text summarization branches out} (pp. 74-81).

Liu, Y., Bhandari, S., \& Pardos, Z. A. (2025). Leveraging LLM respondents for item evaluation: A psychometric analysis. \emph{British Journal of Educational Technology, 56}(3), 1028-1052.

Milano, N., Luongo, M., Ponticorvo, M., \& Marocco, D. (2025). Semantic analysis of test items through large language model embeddings predicts a-priori factorial structure of personality tests. \emph{Current Research in Behavioral Sciences, 8}, 100168.

Moore, S., Costello, E., Nguyen, H. A., \& Stamper, J. (2024, July). An automatic question usability evaluation toolkit. In \emph{International Conference on Artificial Intelligence in Education} (pp. 31-46). Cham: Springer Nature Switzerland.

Murtagh, F., \& Contreras, P. (2012). Algorithms for hierarchical clustering: an overview. \emph{Wiley interdisciplinary reviews: data mining and knowledge discovery, 2}(1), 86-97.

Papineni, K., Roukos, S., Ward, T., \& Zhu, W. J. (2002, July). Bleu: a method for automatic evaluation of machine translation. In \emph{Proceedings of the 40th annual meeting of the Association for Computational Linguistics} (pp. 311-318).

Pelánek, R. (2019). Measuring similarity of educational items: An overview. \emph{IEEE Transactions on Learning Technologies, 13}(2), 354-366.

Rihák, J., \& Pelánek, R. (2017). Measuring Similarity of Educational Items Using Data on Learners\textquotesingle{} Performance. In \emph{International Educational Data Mining Society}. Wuhan, China.

Rodriguez-Torrealba, R., Garcia-Lopez, E., \& Garcia-Cabot, A. (2022). End-to-end generation of multiple-choice questions using text-to-text transfer transformer models. \emph{Expert Systems with Applications, 208}, 118258.

Roju, N. S., Van der Linden, W. J., \& Fleer, P. F. (1995). IRT-based internal measures of differential functioning of items and tests. \emph{Applied Psychological Measurement, 19}(4), 353-368.

Sayin, A., \& Gierl, M. (2024). Using OpenAI GPT to generate reading comprehension items. \emph{Educational Measurement: Issues and Practice, 43}(1), 5-18.

Song, Y., Du, J., \& Zheng, Q. (2025). Automatic item generation for educational assessments: a systematic literature review. \emph{Interactive Learning Environments}, \emph{33}(9), 5386--5405.

Sonnleitner, P., Bernard, S., Michels, M. A., Inostroza-Fernandez, P., Keller, U., Gierl, M. J., Cardoso-Leite, P., \& Hornung, C. (2025). Establishing Cognitive Item Models for Fair and Theory-Grounded Automatic Item Generation: A Large-Scale Assessment Study with Image-Based Math Items. \emph{Applied Measurement in Education, 38}(2), 1-23.

Stanton, J. M., Sinar, E. F., Balzer, W. K., \& Smith, P. C. (2002). Issues and strategies for reducing the length of self-report scales. \emph{Personnel Psychology, 55}(1), 167--194.

Steck, H., Ekanadham, C., \& Kallus, N. (2024, May). Is cosine-similarity of embeddings really about similarity?. In \emph{Companion Proceedings of the ACM Web Conference 2024 (pp. 887-890).}

Tao, L., Cao, J., \& Liu, F. (2017). Quantifying textual terms of items for similarity measurement. \emph{Information Sciences, 415}, 269-282.

Wang, Y., Huang, J., Du, L., Guo, Y., Liu, Y., \& Wang, R. (2025). Evaluating large language models as raters in large-scale writing assessments: A psychometric framework for reliability and validity. \emph{Computers and Education: Artificial Intelligence,} 100481.

Weinberger, A. H., Darkes, J., Del Boca, F. K., Greenbaum, P. E., \& Goldman, M. S. (2006). Items as context: Effects of item order and ambiguity on factor structure. \emph{Basic and Applied Social Psychology, 28}(1), 17-26.

Welbers, K., Van Atteveldt, W., \& Benoit, K. (2017). Text analysis in R. \emph{Communication methods and measures, 11}(4), 245-265.

Wise, S. L., \& Kingsbury, G. G. (2022). Performance decline as an indicator of generalized test-taking disengagement. \emph{Applied Measurement in Education, 35}(4), 272-286.

Xu, S., Wu, Z., Zhao, H., Shu, P., Liu, Z., Liao, W., Li, S., Sikora, A., Liu, T., \& Li, X. (2024). Reasoning before comparison: LLM-enhanced semantic similarity metrics for domain specialized text analysis. \emph{arXiv preprint arXiv:2402.11398.}

Zheng, C., Liu, J., Li, Y., Xu, P., Zhang, B., Wei, R., Zhang, W., Liu, B., \& Huang, J. (2024). A 2PLM-RANK multidimensional forced-choice model and its fast estimation algorithm. \emph{Behavior Research Methods, 56}(6), 6363-6388.

Zhou, K., Ethayarajh, K., Card, D., \& Jurafsky, D. (2022). Problems with cosine as a measure of embedding similarity for high frequency words. \emph{arXiv preprint arXiv:2205.05092.}

\textbf{Appendix}

Table I. List of Items of ECR

{\def\LTcaptype{none} % do not increment counter
\begin{longtable}[]{@{}
  >{\raggedright\arraybackslash}p{(\linewidth - 0\tabcolsep) * \real{1.0000}}@{}}
\toprule\noalign{}
\endfirsthead
\endhead
\bottomrule\noalign{}
\endlastfoot
1. I prefer not to show a partner how I feel deep down. \\
2. I worry about being abandoned. \\
3. I am very comfortable being close to romantic partners. \\
4. I worry a lot about my relationships. \\
5. Just when my partner starts to get close to me, I find myself pulling away. \\
6. I worry that romantic partners won't care about me as much as I care about them. \\
7. I get uncomfortable when a romantic partner wants to be very close. \\
8. I worry a fair amount about losing my partner. \\
9. I don't feel comfortable opening up to romantic partners. \\
10. I often wish that my partner's feelings for me were as strong as my feelings for him/her. \\
11. I want to get close to my partner, but I keep pulling back. \\
12. I often want to merge completely with romantic partners, and this sometimes scares \\
them away. \\
13. I am nervous when partners get too close to me. \\
14. I worry about being alone. \\
15. I feel comfortable sharing my private thoughts and feelings with my partner. \\
16. My desire to be very close sometimes scares people away. \\
17. I try to avoid getting too close to my partner. \\
18. I need a lot of reassurance that I am loved by my partner. \\
19. I find it relatively easy to get close to my partner. \\
20. Sometimes I feel that I force my partners to show more feeling, more commitment. \\
21. I find it difficult to allow myself to depend on romantic partners. \\
22. I do not often worry about being abandoned. \\
23. I prefer not to be too close to romantic partners. \\
24. If I can't get my partner to show interest in me, I get upset or angry. \\
25. I tell my partner just about everything. \\
26. I find that my partner(s) don't want to get as close as I would like. \\
27. I usually discuss my problems and concerns with my partner. \\
28. When I'm not involved in a relationship, I feel somewhat anxious and insecure. \\
29. I feel comfortable depending on romantic partners. \\
30. I get frustrated when my partner is not around as much as I would like. \\
31. I don't mind asking romantic partners for comfort, advice, or help. \\
32. I get frustrated if romantic partners are not available when I need them. \\
33. It helps to turn to my romantic partner in times of need. \\
34. When romantic partners disapprove of me, I feel really bad about myself. \\
35. I turn to my partner for many things, including comfort and reassurance. \\
36. I resent it when my partner spends time away from me. \\
\end{longtable}
}

Table II. Item Parameters and Clustering Results

{\def\LTcaptype{none} % do not increment counter
\begin{longtable}[]{@{}lrrrrrccc@{}}
\toprule\noalign{}
\multirow{2}{*}{item\_id} & \multirow{2}{*}{a} & \multirow{2}{*}{b1} & \multirow{2}{*}{b2} & \multirow{2}{*}{b3} & \multirow{2}{*}{b4} & \multicolumn{3}{c}{Clusters} \\
\cmidrule(lr){7-9}
& & & & & & LLM & BLEU & cosine \\
\midrule\noalign{}
\endhead
\bottomrule\noalign{}
\endlastfoot
Q1 & 1.69 & -1.143 & 0.18 & 0.616 & 2.041 & 3 & 4 & 3 \\
Q2 & 0.363 & -6.227 & -2.669 & -0.986 & 3.267 & 2 & 1 & 1 \\
Q3 & 2.086 & -0.647 & 0.52 & 1.07 & 2.122 & 4 & 3 & 2 \\
Q4 & 0.377 & -6.93 & -2.55 & -0.212 & 4.091 & 2 & 4 & 4 \\
Q5 & 2.255 & -0.953 & 0.138 & 0.644 & 1.704 & 3 & 4 & 4 \\
Q6 & 0.465 & -4.841 & -1.853 & -0.624 & 2.776 & 1 & 4 & 3 \\
Q7 & 2.477 & -0.744 & 0.334 & 0.789 & 1.748 & 3 & 4 & 2 \\
Q8 & 0.307 & -6.862 & -1.972 & 0.583 & 5.836 & 2 & 4 & 4 \\
Q9 & 2.672 & -0.905 & 0.238 & 0.691 & 1.726 & 3 & 3 & 3 \\
Q10 & 0.306 & -6.752 & -2.275 & 0.745 & 4.755 & 1 & 4 & 4 \\
Q11 & 2.206 & -0.974 & 0.143 & 0.661 & 1.736 & 3 & 2 & 2 \\
Q12 & 0.1 & -9.991 & 3.906 & 10 & 10 & 1 & 4 & 2 \\
Q13 & 2.644 & -0.876 & 0.269 & 0.711 & 1.824 & 3 & 3 & 2 \\
Q14 & 0.165 & -10 & -3.141 & 0.259 & 8.3 & 2 & 1 & 1 \\
Q15 & 1.713 & -0.944 & 0.475 & 1.021 & 2.111 & 4 & 2 & 2 \\
Q16 & 0.1 & -10 & 4.565 & 10 & 10 & 1 & 4 & 2 \\
Q17 & 2.658 & -0.873 & 0.317 & 0.875 & 2.004 & 3 & 3 & 2 \\
Q18 & 0.14 & -10 & -6.576 & -1.353 & 9.218 & 1 & 4 & 4 \\
Q19 & 1.904 & -1.35 & 0.204 & 0.929 & 2.165 & 4 & 2 & 2 \\
Q20 & 0.1 & -10 & -1.349 & 5.494 & 10 & 1 & 4 & 3 \\
Q21 & 1.493 & -1.925 & -0.515 & 0.1 & 1.428 & 3 & 4 & 2 \\
Q22 & 0.264 & -7.871 & -2.005 & 0.414 & 6.172 & 2 & 1 & 1 \\
Q23 & 2.349 & -0.807 & 0.515 & 1.113 & 2.2 & 3 & 3 & 2 \\
Q24 & 0.112 & -10 & -6.058 & 0.855 & 10 & 1 & 4 & 3 \\
Q25 & 1.649 & -1.182 & 0.184 & 0.751 & 1.912 & 4 & 4 & 4 \\
Q26 & 0.266 & -5.985 & 0.583 & 4.175 & 10 & 1 & 4 & 4 \\
Q27 & 1.787 & -0.976 & 0.676 & 1.289 & 2.401 & 4 & 4 & 4 \\
Q28 & 0.1 & -10 & 1.431 & 8.671 & 10 & 2 & 4 & 3 \\
Q29 & 1.369 & -2.053 & -0.286 & 0.522 & 1.903 & 4 & 4 & 3 \\
Q30 & 0.1 & -10 & -7.434 & 0.616 & 10 & 1 & 4 & 4 \\
Q31 & 1.708 & -0.971 & 0.692 & 1.26 & 2.429 & 4 & 4 & 3 \\
Q32 & 0.1 & -10 & -9.853 & -1.631 & 10 & 1 & 4 & 3 \\
Q33 & 1.501 & -0.997 & 0.975 & 1.878 & 2.962 & 4 & 4 & 2 \\
Q34 & 0.1 & -10 & -10 & -4.365 & 10 & 2 & 4 & 3 \\
Q35 & 0.1 & -10 & -10 & -9.325 & 10 & 4 & 4 & 4 \\
Q36 & 0.104 & -10 & 2.49 & 10 & 10 & 1 & 4 & 4 \\
\end{longtable}
}

\includegraphics[width=4.86614in,height=2.5043in,alt={A diagram of a graph AI-generated content may be incorrect.}]{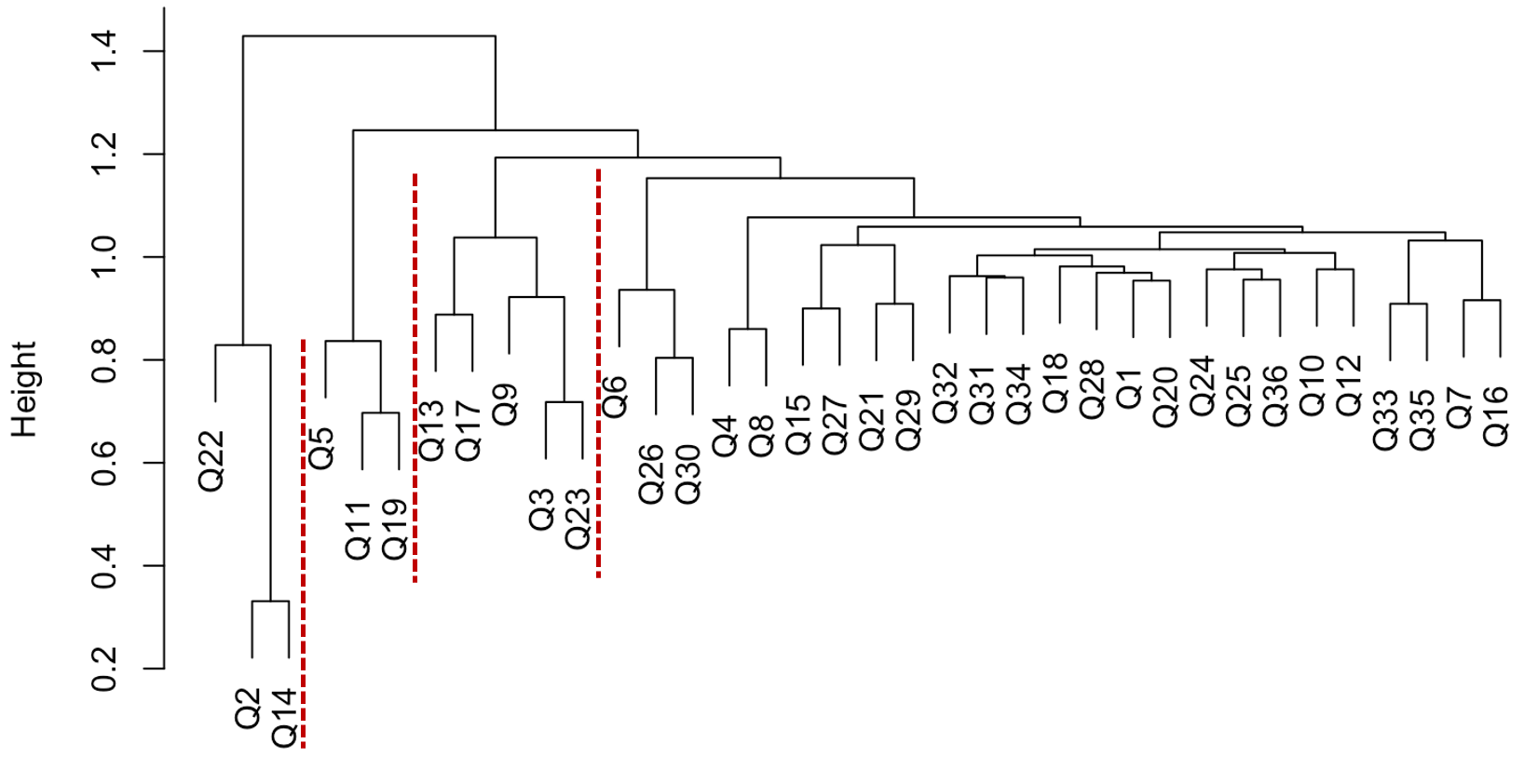}

Figure I. Hierarchical cluster results for BLEU similarity.

\includegraphics[width=4.66929in,height=2.38004in,alt={A diagram of a structure AI-generated content may be incorrect.}]{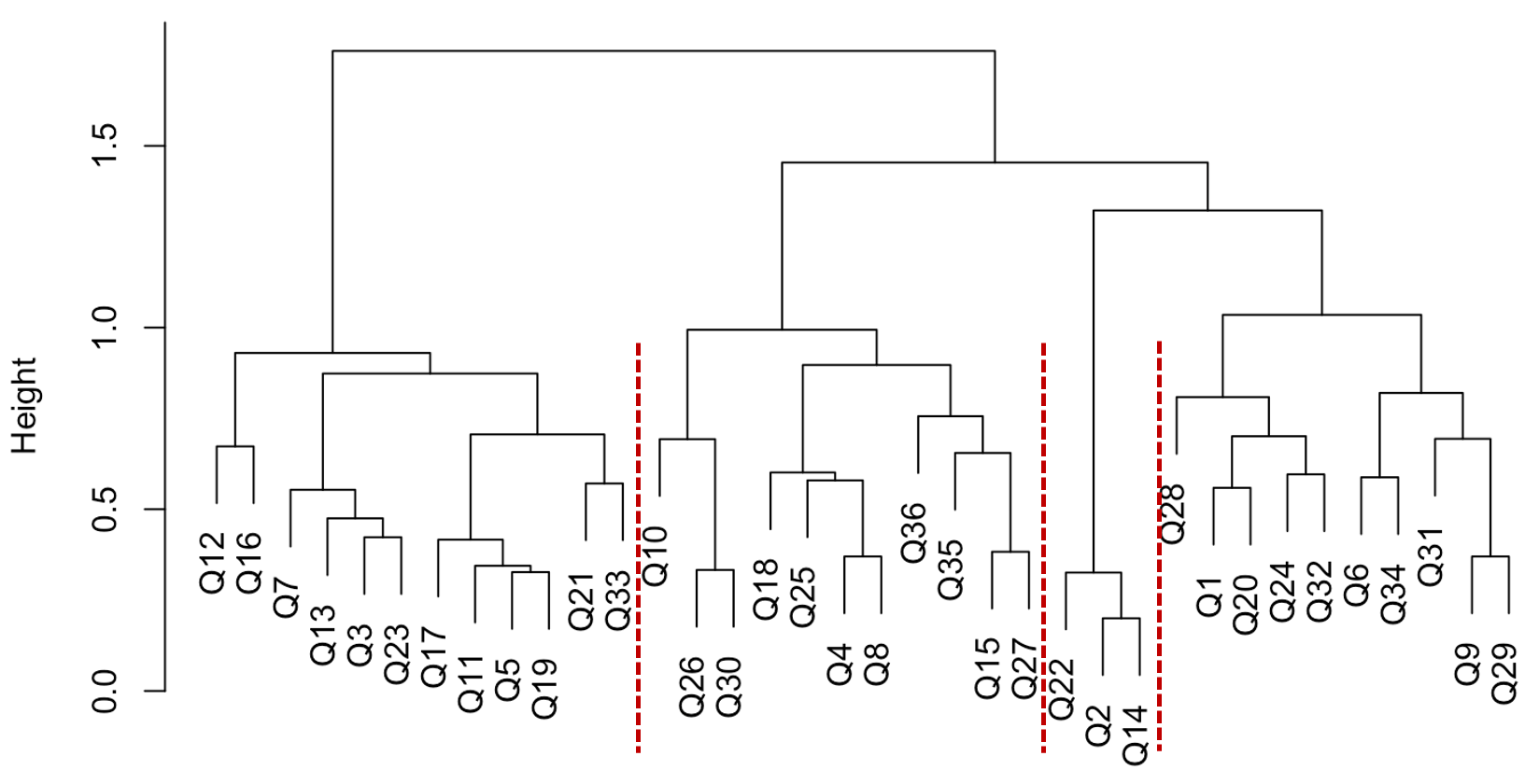}

Figure II. Hierarchical cluster results for cosine similarity.

\end{document}